\pdfoutput=1
\documentclass[11pt]{article}

\usepackage{arxiv}

\usepackage{times}
\usepackage{latexsym}

\usepackage{amsmath}
\usepackage{url}
\usepackage{natbib}
\usepackage{graphicx}
\usepackage[utf8]{inputenc} % allow utf-8 input
\usepackage[T1]{fontenc}    % use 8-bit T1 fonts
\usepackage{url}            % simple URL typesetting
\usepackage{booktabs}       % professional-quality tables
\usepackage{amsfonts}       % blackboard math symbols
\usepackage{nicefrac}       % compact symbols for 1/2, etc.
\usepackage{microtype}      % microtypography
\usepackage{changepage}
\usepackage{amssymb}
\usepackage{graphicx}%
\usepackage{xcolor}
\usepackage{url}
\usepackage{wrapfig}
\usepackage{graphicx}
\usepackage{amsthm}
\usepackage{amssymb}
\usepackage{subcaption}
\usepackage{threeparttable}
\usepackage{pifont}% http://ctan.org/pkg/pifont
\usepackage[T1]{fontenc}
\usepackage{scrextend}
\usepackage[utf8]{inputenc}

\usepackage{microtype}
\usepackage{rotating}
\usepackage{multirow}
\usepackage[bottom]{footmisc}
\title{LazyFormer: Self Attention with Lazy Update}
\author{
  Chengxuan Ying \\
  Dalian University of Technology \\
  \texttt{yingchengsyuan@gmail.com} \\
\And
  Guolin Ke \\
  Microsoft Research \\
  \texttt{guolin.ke@microsoft.com} \\
\AND
  Di He \\
  Microsoft Research \\
  \texttt{dihe@microsoft.com} \\
\And
  Tie-Yan Liu \\
  Microsoft Research \\
  \texttt{tyliu@microsoft.com} \\
}

\begin{document}
\maketitle

\begin{abstract}

Improving the efficiency of Transformer-based language pre-training is an important task in NLP, especially for the self-attention module, which is computationally expensive. In this paper, we propose a simple but effective solution, called \emph{LazyFormer}, which computes the self-attention distribution infrequently. LazyFormer composes of multiple lazy blocks, each of which contains multiple Transformer layers. In each lazy block, the self-attention distribution is only computed once in the first layer and then is reused in all upper layers.  In this way, the cost of computation could be largely saved. We also provide several training tricks for LazyFormer. Extensive experiments demonstrate the effectiveness of the proposed method.

\end{abstract}

\section{Introduction}

Using pre-trained contextual representations (e.g., BERT) \cite{devlin2018bert} have become the standard way to improve the performance on the downstream tasks in natural language processing. Transformer \cite{vaswani2017attention} is the basic building block for almost all pre-training methods \cite{liu2019roberta, clark2019electra, devlin2018bert}. A Transformer layer is composed of an efficient densely connected network operated on each position separately and a less-efficient self-attention module, which costs $O(n^2)$ ($n$ is sequence length) time and space. This quadratic cost becomes a bottleneck in Transformer, especially when $n$ is large. Many recent works \cite{wang2020linformer, beltagy2020longformer, zaheer2020big, kitaev2020reformer, choromanski2020rethinking} tried to reduce the $O(n^2)$ cost to $O(n\sqrt{n})$ or $O(n \text{log} n)$, by sparsifying or approximating the attention matrix. 

\begin{figure}[t]
  \centering
  \includegraphics[width=0.55\textwidth]{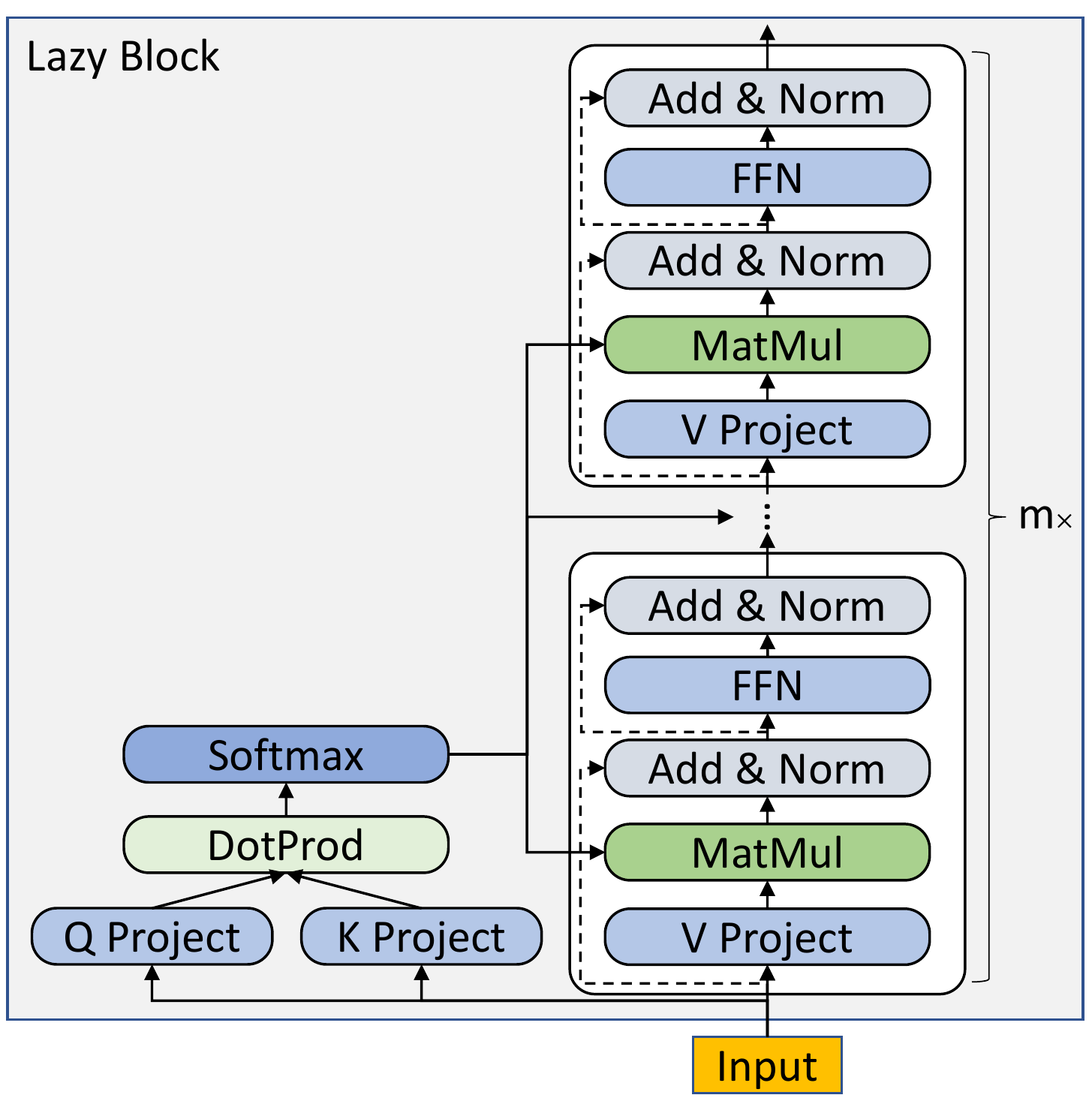}
%   %   \vspace{-20pt}
  \caption{The basic block in LazyFormer.}
  \label{fig:arch}
%   %   \vspace{-20pt}
\end{figure}

In this paper, different from previous works, we explore a simple idea to improve the model efficiency by computing the self-attention \emph{infrequently}. We call it \emph{LazyFormer}. More specifically, a LazyFormer consists of multiple basic blocks, and each basic block is composed of $m$ Transformer layers, as shown in Figure \ref{fig:arch}. In each block, we only compute Query-Key dot-product attention once in the first layer and then reuse it in the upper layers. In this way, LazyFormer only needs to calculate the attention once in $m$ Transformer layers, reducing the computational cost from $O(n^2)$ to $O(n^2/m)$.

We conduct extensive experiments to verify the efficiency and effectiveness of LazyFormer in language pre-training. From the results, we observe: 1) compared with a standard Transformer model of the same model capacity, LazyFormer is 1.3x faster without hurting any performance. 2) LazyFormer allows us to train the larger models effectively. Specifically, with the same pre-training cost, the larger LazyFormer can outperform the baseline by 1 point. Besides, it can achieve a better GLUE score by only using 50\% costs.
% It achieves a better GLUE score by only using 50\% pre-training costs and outperforms the baseline by 1 point. 
3) LazyFormer can better handle longer sequences. With $n$ increasing, the speed-up is more significant.

% %   \vspace{-5pt}
\paragraph{Related work}
LazyFormer is inspired by several recent works which investigate what self-attention module learns in each layer \cite{clark2019does, vig2019analyzing, vig2019multiscale, xiao2019sharing}. From these previous works, we can easily see that the distributions of self-attention outputs are similar between the adjacent Transformer layers. Such observations motivate us to reuse the self-attention outputs in the lower layers to the upper layers. 

There are some works that leverage such observation for better training, like \cite{gong2019efficient} and \cite{lan2019albert}. However, these works share the parameters of self-attention in different layers but still need to compute self-attention in all layers. Therefore they cannot save the computational costs. Our work is also orthogonal to the works that modify the self-attention modules to reduce the cost, such as Reformer\cite{kitaev2020reformer} and Linformer\cite{wang2020linformer}. These works aim to reduce the computation in each layer, while ours is to reuse the self-attention outputs from lower layers. Both works could be combined and further reduce the cost of language pre-training.

\section{LazyFormer}

The attention module \citep{vaswani2017attention} in Transformer can be generally formulated as querying a dictionary with key-value pairs, e.g., $\text{Attention}(Q,K,V)=\text{softmax}(\frac{QK^T}{\sqrt{d}})V$, where $d$ is the dimensionality of the hidden representations. In the self-attention module, $Q$ (Query), $K$ (Key) and $V$ (Value) are parameterized from the input $x$, i.e., $\text{Attention}(xW^Q,xW^K,xW^V)$, where $W^Q$, $W^K$ and $W^V$ are the projection matrices. It is easy to see that the computational complexity of $\text{softmax}(\frac{QK^T}{\sqrt{d}})$ is $O(n^2)$, since it has to compute the pair-wise correlations for all input tokens, and produces an $n \times n$ matrix (attention matrix). When we use stacked $k$ Transformer layers, as all attention matrices need to be computed, the total cost of self-attention calculation is $O(kn^2)$.

Many recent works \cite{vig2019analyzing, vig2019multiscale} show the attention matrices are similar in different layers, especially in the adjacent layers. Therefore, we argue that the attention matrix maybe does not need to be computed in every layer. For example, we can only calculate the attentions in a layer and reuse it for multiple adjacent upper layers. Formally, we define a new Transformer variant called \emph{LazyFormer}, which composes of several ``lazy'' blocks. Each lazy block consists of $m$ Transformer layers, as shown in Figure \ref{fig:arch}. In each block, the attention matrix will be only computed in the first layer using the input to the block, and then reused by all $m-1$ upper layers. We can stack $k/m$ lazy blocks to construct a $k$-layer Transformer model. In this way, the cost of computing the attention could be reduced from $O(kn^2)$ to $O(kn^2/m)$. 

%For simplicity, LazyFormer uses uniform blocks, in which the numbers of layers are the same across blocks, like M3x4. One can also use the non-uniform blocks, like M2M4M6 (3 blocks, each with 2, 4, and 6 layers respectively). Our ablation studies (see Experiment) show uniform blocks and non-uniform blocks achieve a similar performance when using the same number of layers.

Based on the design of lazy block, we further use the following two additional methods to improve LazyFormer:

% %   \vspace{-8pt}
\paragraph{Wider Layers.} In LazyFormer, the number of projection matrices $W^Q$ and $W^K$ are reduced. Therefore, the total parameters in LazyFormer are less than the stacked Transformers when using the same width and depth. For example, the BERT-based model has 12 layers. The embedding dimension is set to 768, and the hidden dimension is set to 3072. This configuration leads to a model size of about 110M parameters. If we use $m=3$ for LazyFormer in the same setting, the model only contains about 100M parameters. 

As the model capacity plays an important role in language pre-training \cite{raffel2019exploring}, we can slightly increase the hidden/embedding dimension in LazyFormer, to match the same number of parameters. Note that increasing the hidden/embedding dimension only slightly affects the efficiency. Wider LazyFormer is still much faster than a Transformer of the same depth. Besides, we can even increase the width of LazyFormer until its forward/backward speed match the Transformer to achieve better performance when using the same pre-training cost. 

% %   \vspace{-8pt}
\paragraph{Remove dropout in self-attention.} Dropout is used in self-attention by default. The cost of that dropout is also $O(n^2)$, as it is applied on the $n \times n$ attention matrix. Recent work \cite{lan2019albert} shows the dropout in self-attention can be safely removed, without hurting the performance. Therefore, for better efficiency, we also remove the dropout in self-attention in LazyFormer.

\section{Experiment}
To verify the performance of the proposed LazyFormer, we conduct extensive experiments and demonstrate the results in this section. We use BERT-Base (112M parameters) architecture for all experiments. Specifically, BERT-Base is consists of 12 Transformer layers, in which the embedding dimension is 768, the number of attention heads is 12, and the hidden dimension is 3072. Besides absolute positional encoding, we further use the relative positional encoding \cite{raffel2019exploring} in the self-attention module for better performance. We provide all the experimental details and results in the Appendix.

% %   \vspace{-5pt}
\subsection{Experimental Design\label{sec:exprimentdesign}}

Following \citet{devlin2018bert}, we use the 16GB corpus (English Wikipedia corpus and BookCorpus \citep{moviebook}) for pre-training. We set the vocabulary size (sub-word tokens) as 32,768, sequence length as 512, and batch size as 256. We use the GLUE (\textbf{G}eneral \textbf{L}anguage \textbf{U}nderstanding \textbf{E}valuation) dataset \citep{DBLP:journals/corr/abs-1804-07461} as the downstream tasks to evaluate the performance of the pre-trained models. All codes are implemented based on \emph{fairseq} \citep{ott2019fairseq} in \emph{PyTorch} \citep{paszke2017automatic}. All models are run on 8 NVIDIA Tesla V100 GPUs with mixed-precision \citep{micikevicius2017mixed}.

We use M$\beta$x$\gamma$ to denote LazyFormer structure, where $\beta$ is the number of layers in each lazy block, and $\gamma$ is number of total blocks. For example, M2x6 denotes the LazyFormer with 6 blocks, each with 2 Transformer layers.

% %   \vspace{-5pt}
\subsection{Overall Comparison}

\begin{table*}[t]
\centering
\caption{GLUE scores on dev set. All models are pre-trained by 16GB data. Both M2x6 and M2x6-S are the LazyFormer with 6 two-layered blocks, without dropout in self-attention. M2x6-S uses the same parameter size as BERT. For the ablation study, based on M2x6-S, M2x6-SD keeps the dropout in self-attention. M2x6 increases the parameter to match the same pre-training cost as BERT. M2x6$^{mid}$ is the intermediate 500k-step checkpoint M2x6. The details of models are shown in Table \ref{tab:overall_meta}.}
% %   \vspace{-10pt}
\label{tab:overall}
\begin{tabular}{lcccccccccccc}
\toprule
 & Steps & MNLI\footnotesize{-m/mm} & QNLI & QQP & SST & CoLA & MRPC & RTE & STS & Avg.  \\
\hline
BERT&1$M$ & \textbf{85.72/85.67} & \textbf{92.07} & 91.21 & 92.78 & \textbf{58.80} & \textbf{88.73} & 67.51 & 89.23 & 83.52 \\
M2x6-S &1$M$ & 85.69/85.46 & 91.14 & 91.13 & 92.66 & 57.86 & 87.25 & \textbf{72.20} & \textbf{89.79} & \textbf{83.69} \\
M2x6-SD &1$M$ & 85.34/85.50 & 91.65 & \textbf{91.22} & \textbf{93.23} & 57.85 & 86.76 & 70.40 & 89.69 & 83.52  \\
M2x6 &1$M$   & \textbf{86.34/86.31} & \textbf{91.82} & \textbf{91.38} & \textbf{93.00} & \textbf{60.77} & \textbf{87.75} & \textbf{73.65} & \textbf{90.00} & \textbf{84.56} \\
\midrule
M2x6$^{mid}$ &500$k$ & 85.81/85.58 & 91.62 & 91.31 & 92.66 & 58.81 & 87.75 & 71.48 & 88.90 & 83.77 \\
\bottomrule
\end{tabular}
% %   \vspace{-8pt}
\end{table*}

\begin{figure*}[b]
\begin{subfigure}{.33\textwidth}
  \centering
  \includegraphics[width=2.1in]{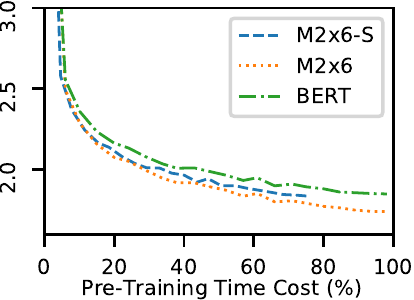}  
  \caption{Validation loss in pre-training.}
  \label{fig:exp-all-valid}
\end{subfigure}
\begin{subfigure}{.33\textwidth}
  \centering
  \includegraphics[width=2.1in]{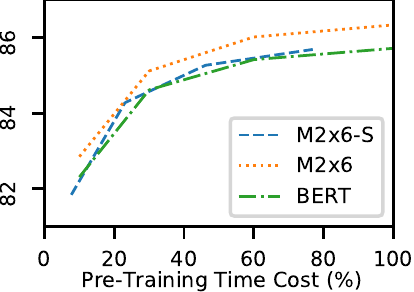}  
  \caption{MNLI-m score.}
  \label{fig:exp-all-mnli}
\end{subfigure}
\begin{subfigure}{.33\textwidth}
  \centering
  \includegraphics[width=2.1in]{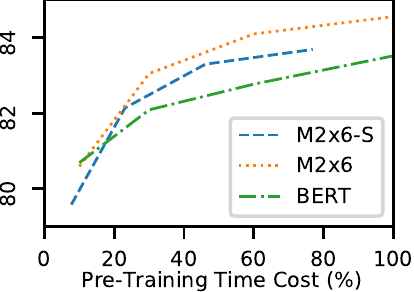}  
  \caption{GLUE average score.}
  \label{fig:exp-all-glue}
\end{subfigure}
% %   \vspace{-10pt}
\caption{Both M2x6-S and M2x6 converge much faster than the baselines. Besides, M2x6 achieves better performance in downstream tasks while using much fewer pre-training costs.}
% %   \vspace{-15pt}
\label{fig:exp-overall}
\end{figure*}

\begin{table}[t]
\centering
\caption{Details of models in Table \ref{tab:overall}. $W$ is the hidden dimension in feed-forward layers, $H$ is the embedding dimension, and $N$ is the number of attention heads.}
% %   \vspace{-10pt}

\label{tab:overall_meta}
\begin{tabular}{lcccc}
\toprule
 & Params & ($W$,$H$,$N$) & Time \protect\footnotemark & Speedup \\
\hline
BERT & 112$M$ & (3072,768,12) & 45 & 1.0x \\ 
M2x6-S& 112$M$ & (3456,768,12) & 35 & 1.3x \\
M2x6-SD & 112$M$ & (3456,768,12) & 38 & 1.2x \\
M2x6 & 157$M$ & (4480,896,14) & 45 & 1.0x \\
% \midrule
% \multicolumn{5}{c}{LazyFormer w/ Dropout} \\
% M1x12& 112$M$ & (3072,768,12) & 40 & 1.0x \\
% M2x6 & 112$M$ & (3456,768,12) & 35 & 1.1x \\
% M3x4 & 112$M$ & (3584,768,12) & 34 & 1.2x \\
% M4x3 & 112$M$ & (3648,768,12) & 33 & 1.2x \\
% M6x2 & 112$M$ & (3712,768,12) & 32 & 1.3x \\
\bottomrule
\end{tabular}
% %   \vspace{-10pt}
\end{table}

\begin{table}[h]
\centering
\caption{Comparison for different LazyFormer layouts. To match the same parameter size as BERT, we increase the hidden dimension $W$ for different layouts. We keep the dropout in self-attention in all settings.}
%   \vspace{-10pt}
\label{tab:layer}
\begin{tabular}{lccc}
\toprule
  & $W$ & Time \footref{footnote1} & GLUE-Avg. \\
\hline
BERT (M1x12-SD)& 3072 & 45 & \textbf{83.52} \\
M2x6-SD & 3456 & 38 & \textbf{83.52} \\
M3x4-SD & 3584 & 35 & 82.94 \\
M4x3-SD & 3648 & 34 & 83.22 \\
M6x2-SD & 3712 & 33 & 82.47 \\
\midrule
M5M3M2M2-SD & 3584 & 35 & 82.89 \\
M2M2M3M5-SD & 3584 & 35 & 82.82 \\
\bottomrule
\end{tabular}
%   \vspace{-15pt}
\end{table}

First, we set up models for the overall comparison. Besides the baseline BERT model, we set up two LazyFormer variants: 1) \textbf{M2x6-S}, which uses six lazy blocks with two Transformer layers in each block, and increases hidden dimension to 3456, to retain the same parameter size as BERT; 2) \textbf{M2x6}, which increases hidden dimension to 4480, embedding dimension to 896, and the number of attention heads to 14, to retain the same pre-training cost as BERT. Both M2x6 and M2x6-S remove dropout in the self-attention module. The detailed settings can be found in Table~\ref{tab:overall_meta}. 

The results are shown in Table~\ref{tab:overall}. Firstly, M2x6-S achieves a slight improvement over BERT but is about 1.3x faster. This result indicates that LazyFormer is much more efficient, without hurting any performance. 

Furthermore, M2x6 outperforms baselines by 1 point in terms of GLUE average score and is consistently better on almost all GLUE tasks. This demonstrates another important strength of LazyFormer: it allows us to increase the model capacity to achieve better performance while using the same computational cost. Besides, from Table 2, we can also see that the results of M2x6$^{mid}$ (intermediate 500k-step checkpoint of M2x6) are already competitive to that of  BERT (trained for 1M steps). This suggests LazyFormer can use a significantly short time to learn a better model. As shown in Figure~~\ref{fig:exp-overall}, both M2x6 and M2x6-S converge much faster than BERT in terms of both pre-training validation loss and down-stream task performance. 

%As a summary, the experimental results show the effectiveness and efficiency of the proposed LazyFormer. In the following subsection, we will examine each modification in LazyFormer to check whether it is useful.  

\footnotetext{Time indicates the training wall time of 100 iterations by 8 V100 GPUs. \label{footnote1}}

%   \vspace{-5pt}
\subsection{Ablation Studies}

%   \vspace{-5pt}
\paragraph{Remove dropout in self-attention.} As aforementioned, dropout in self-attention brings additional costs. So we empirically study whether using dropout is essential. First, as shown in Table \ref{tab:overall}, compared M2x6-S with M2x6-SD (M2x6-S with dropout in self-attention), removing dropout in self-attention slightly improves the performance. Besides, as shown in Table \ref{tab:overall_meta}, removing dropout in self-attention can bring the 10\% speed-up. In short, removing dropout in self-attention can improve efficiency without hurting the performance.

%   \vspace{-5pt}
\paragraph{Different layouts.} There are many design choices in LazyFormer. For example, one can set low-level blocks with more/fewer layers or set the whole model with more/fewer blocks. We study how different layouts perform and summarize the results in Table~\ref{tab:layer}. 
First, we find that when the number of blocks decreases, the training is faster, but the acceleration rate is not significant. At the same time, the final performance gets worse. Therefore, we observe using six blocks is a trade-off choice for both efficiency and effectiveness. 
Second, we find the number of blocks is the key factor for the performance. For all the 4-block models, setting each block with 3 layers (M3x4-SD) achieves similar performance to the models that set different numbers of layers in different blocks (M5M3M2M2-SD \footnote{M5M3M2M2 denotes the model with 4 sequential lazy blocks, each with 5, 3, 2, 2 layers respectively.}, M2M2M3M5-SD). 

\begin{table}[t]
\centering
\caption{Comparison for different ($W$, $H$, $N$) settings under the same pre-training cost.}
%   \vspace{-10pt}
\label{tab:vary}
\begin{tabular}{lcccc}
\toprule
 & Params & ($W$,$H$,$N$) & MNLI\footnotesize{-m/mm} & GLUE-Avg. \\
\hline
M2x6-S& 112$M$ & (3456,768,12) & 85.72/85.67 & 83.69 \\
M2x6  & 157$M$ & (4480,896,14) & \textbf{86.34/86.31} & 84.56 \\
M2x6  & 161$M$ & (6144,768,12) & \textbf{86.08/86.57} & \textbf{84.77} \\
M2x7  & 148$M$ & (4480,768,12) & 86.25/86.26 & 84.75 \\
\bottomrule
\end{tabular}
%   \vspace{-10pt}
\end{table}

\begin{figure}[t]
  %   \vspace{+5pt}
  \centering
  \includegraphics[width=0.60\textwidth]{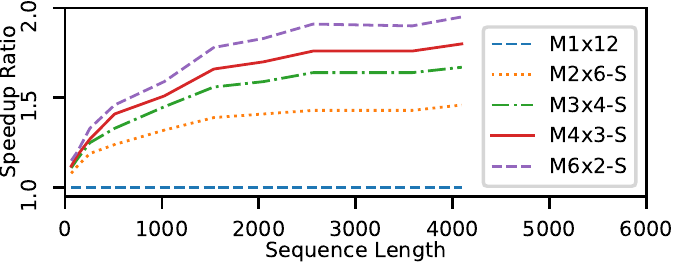}  
  %   \vspace{-20pt}
  \hspace{+14pt}
  \caption{Speedup ratio of \textit{LazyFormer} under different settings of lazy blocks. }
  \label{speedup}
  %   \vspace{-20pt}
\end{figure}

%   \vspace{-5pt}
\paragraph{Different model settings under the same computational cost.} As aforementioned, LazyFormer allows us to use a larger model capacity within the same computational cost. Therefore, we investigate different settings to increase model parameters and summarize the results in Table~\ref{tab:vary}. Specifically, we study the performance when we increase hidden dimension $W$, embedding dimension $H$, the number of attention heads $N$, or stack more blocks. From the results, we find all these settings can achieve much better performance than the baseline.

%   \vspace{-5pt}
\paragraph{Speedup for longer sequences.} If considering the cost from densely connected layer, LazyFormer reduces the cost from $O(n^2 + p)$ to $O(n^2/m + p)$, where $p$ is the cost of densely connected layer. Therefore, with $n$ increasing, the cost is dominated by the $n^2$ term, and the speedup brought by LazyFormer will be also larger. To study this, we provide speedup ratios given different $n$, shown in Figure~\ref{speedup}. It is easy to find the LazyFormer can bring almost 2x speed-up for long sequences.

\section{Conclusion}
%   \vspace{-6pt}
We propose LazyFormer, which lazily updates the self-attention module and lead to an efficient model architecture of the Transformer. Specifically, LazyFormer composes of multiple lazy blocks, each of which contains multiple Transformer layers. In each lazy block, the self-attention distribution is only computed once in the first layer and then is reused in all upper layers. Besides, LazyFormer further removes dropout in the self-attention module for better efficiency and increases model capacity for better performance. Extensive experimental results demonstrate both the efficiency and effectiveness of LazyFormer.

\clearpage

\bibliography{anthology,custom}
\bibliographystyle{acl_natbib}

\clearpage

\appendix

%
% \section{Example Appendix}
% \label{sec:appendix}

\section{Experimental Details}
\paragraph{Pre-training.} 
Following BERT \citep{devlin2018bert}, we use both English Wikipedia corpus and BookCorpus \citep{moviebook} for language pre-training. By concatenating these two datasets, we can obtain a corpus with 16GB raw text. We adopt the following consecutive pre-processing steps: segmenting documents into sentences by Spacy\footnote{\url{https://spacy.io}}, normalizing, lower-casing, and tokenizing the texts by Moses decoder \citep{Koehn2007MosesOS}, and finally, applying byte pair encoding (BPE) \citep{DBLP:journals/corr/SennrichHB15} setting the vocabulary size to 32,768. 

We found the data cleaning is important for language pre-training. To this end, we de-duplicate the documents, normalize the punctuations, concatenate the short sequences, replace the URL and other hyperlinks to special tokens, and filter the low-frequency tokens. Therefore, our re-implemented baselines, like BERT, can achieve a higher average GLUE scores than the original papers.

We use masked language modeling as the objective of pre-training. We remove the next sentence prediction task and use \textit{FULL-SENTENCES} mode to pack sentences as suggested in RoBERTa \citep{liu2019roberta}. We train the models for 1000$k$ steps where the batch size is 256 and the maximum sequence length is 512. The masked probability is set to 0.15, with replacing 80\% of the masked positions by \texttt{[MASK]}, 10\% by randomly sampled words, and keep the remaining 10\% unchanged. We use Adam \citep{DBLP:journals/corr/KingmaB14} as the optimizer, and set the its hyperparameter $\epsilon$ to 1e-6 and $(\beta1, \beta2)$ to (0.9, 0.999). The peak learning rate is set to 1e-4 with a 10$k$-step warm-up stage. After the warm-up stage, the learning rate decays linearly to zero. We set the dropout probability to 0.1, gradient clip norm to 1.0, and weight decay to 0.01. Besides the final checkpoint, we also save intermediate checkpoints and fine-tune them on downstream tasks, to check the efficiency of different methods.

\paragraph{Fine-tuning.}
We use the GLUE (\textbf{G}eneral \textbf{L}anguage \textbf{U}nderstanding \textbf{E}valuation) dataset \citep{DBLP:journals/corr/abs-1804-07461} as the downstream tasks to evaluate the performance of the pre-trained models. Specifically, we use nine tasks in GLUE, including CoLA, RTE, MRPC, STS-B, SST, QNLI, QQP, and MNLI-m/mm. For the evaluation metrics, we report Matthews correlation for CoLA, Pearson correlation for STS-B, and accuracy for other tasks. We use the same optimizer (Adam) with the same hyperparameters as in pre-training. Following previous works, we search the learning rates during the fine-tuning for each downstream task. The setting details are listed in Table~\ref{tab:ds_space}. For a fair comparison, we do not apply any tricks for fine-tuning. Each configuration will be run five times with different random seeds, and the \emph{median} of these five results on the development set will be used as the performance of one configuration. We will ultimately report the best number over all configurations. We also provide all the detailed results in Table~\ref{tab:overall}.

\begin{table*}
    \centering 
    \caption{Hyperparameters for the pre-training and fine-tuning. } \label{tab:ds_space}
%\addtolength{\tabcolsep}{-0.5pt}
\begin{threeparttable}
\begin{tabular}{lcc}
\toprule
& Pre-training & Fine-tuning \\ \hline
\textbf{Max Steps} & 1$M$ & - \\
\textbf{Max Epochs} & - & 10 \\ 
\textbf{Learning Rate} & 1e-4 & \{2e-5, 3e-5, 4e-5, 5e-5\}  \\ 
\textbf{Batch Size} & 256 & 32 \\ 
\textbf{Warm-up Ratio} & 0.01 & 0.06 \\ 
\textbf{Sequence Length} & 512 & 512 \\
\textbf{Learning Rate Decay} & Linear & Linear \\ 
\textbf{Adam $\epsilon$} &  1e-6 & 1e-6 \\ 
\textbf{Adam ($\beta_1$, $\beta_2$)} &  (0.9, 0.999) & (0.9, 0.999) \\ 
\textbf{Clip Norm} &  1.0 & 1.0 \\ 
\textbf{Dropout} & 0.1 & 0.1 \\ 
\textbf{Weight Decay} & 0.01 & 0.01 \\ 
\bottomrule
\end{tabular}
%\addtolength{\tabcolsep}{0.5pt}
\end{threeparttable}
\end{table*}

\clearpage

\begin{sidewaystable}
% %   \vspace{+240pt}
\centering
\caption{GLUE scores on dev set. All models are pre-trained by 16GB data. Both task scores and GLUE-average scores evaluated at 100$k$, 300$k$, 600$k$, 1$M$ steps are reported in the table. Models ending with '-S' keeps the same parameter size with BERT, models ending with '-D' removes all dropout layers in self-attention. }
\label{tab:overall}
\addtolength{\tabcolsep}{-2pt}    
\begin{tabular}{l|cccc|cccccccccccc}
\toprule
 & Params & ($W$,$H$,$N$) & Pre-training Time & Speedup Ratio & Steps & MNLI\footnotesize{-m/mm} & QNLI & QQP & SST & CoLA & MRPC & RTE & STS & Avg.  \\
\hline
\multirow{4}{*}{BERT (M1x12-SD)}&\multirow{4}{*}{112$M$}&\multirow{4}{*}{(3072,768,12)}&\multirow{4}{*}{45}&\multirow{4}{*}{1.0x}&100$k$ & 82.31/82.51 & 89.93 & 90.76 & 90.60 & 51.79 & 86.03 & 64.26 & 88.03 & 80.69  \\
&&&&&300$k$ & 84.62/84.70 & 91.43 & 91.09 & 91.86 & 54.69 & 87.50 & 64.62 & 88.26 & 82.09 \\
&&&&&600$k$ & 85.42/85.25 & 91.69 & 91.18 & 92.55 & 55.48 & 88.73 & 65.70 & 88.96 & 82.77  \\
&&&&&1$M$   & 85.72/85.67 & 92.07 & 91.21 & 92.78 & 58.80 & 88.73 & 67.51 & 89.23 & 83.52 \\
\midrule
\multirow{4}{*}{M2x6-S}&\multirow{4}{*}{112$M$}&\multirow{4}{*}{(3456,768,12)}&\multirow{4}{*}{35}&\multirow{4}{*}{1.3x}&100$k$ & 81.84/82.35 & 89.15 & 90.68 & 90.37 & 46.05 & 84.31 & 63.54 & 87.91 & 79.58 \\
&&&&&300$k$ & 84.28/84.05 & 90.77 & 91.03 & 91.97 & 53.18 & 86.52 & 68.23 & 89.18 & 82.14 \\
&&&&&600$k$ & 85.27/85.04 & 91.14 & 91.18 & 92.43 & 57.28 & 87.75 & 70.04 & 89.60 & 83.30 \\
&&&&&1$M$   & 85.69/85.46 & 91.14 & 91.13 & 92.66 & 57.86 & 87.25 & 72.20 & 89.79 & 83.69 \\
\midrule
\multirow{4}{*}{M2x6-SD}&\multirow{4}{*}{112$M$}&\multirow{4}{*}{(3456,768,12)}&\multirow{4}{*}{38}&\multirow{4}{*}{1.2x}&100$k$ & 82.23/82.39 & 89.58 & 90.76 & 90.94 & 51.50 & 84.80 & 64.62 & 87.81 & 80.52 \\
&&&&&300$k$ & 84.74/84.59 & 91.27 & 91.15 & 91.74 & 56.83 & 87.99 & 64.62 & 88.37 & 82.37 \\
&&&&&600$k$ & 85.30/85.25 & 91.49 & 91.22 & 93.00 & 58.80 & 87.75 & 67.87 & 89.19 & 83.32 \\
&&&&&1$M$   & 85.34/85.50 & 91.65 & 91.22 & 93.23 & 57.85 & 86.76 & 70.40 & 89.69 & 83.52 \\
\midrule
\multirow{4}{*}{M2x6}&\multirow{4}{*}{157$M$}&\multirow{4}{*}{(4480,896,14)}&\multirow{4}{*}{45}&\multirow{4}{*}{1.0x}&100$k$ & 82.85/82.92 & 89.64 & 90.86 & 90.48 & 47.48 & 86.52 & 66.07 & 88.49 & 80.59 \\
&&&&&300$k$ & 85.12/84.94 & 91.12 & 91.23 & 92.09 & 56.63 & 87.75 & 69.68 & 88.89 & 83.05 \\
&&&&&600$k$ & 86.02/85.62 & 91.78 & 91.37 & 92.43 & 59.83 & 87.50 & 72.92 & 89.42 & 84.10 \\
&&&&&1$M$   & 86.34/86.31 & 91.82 & 91.38 & 93.00 & 60.77 & 87.75 & 73.65 & 90.00 & 84.56 \\
\bottomrule

% part2
\bottomrule
\multirow{4}{*}{M3x4-SD} &\multirow{4}{*}{112$M$}&\multirow{4}{*}{(3584,768,12)}&\multirow{4}{*}{35}&\multirow{4}{*}{1.3x}&100$k$ & 82.24/82.45 & 89.11 & 90.79 & 90.71 & 48.34 & 85.78 & 66.43 & 88.63 & 80.50 \\
&&&&&300$k$ & 84.26/84.49 & 90.26 & 91.06 & 92.20 & 51.03 & 87.01 & 66.43 & 88.66 & 81.71 \\
&&&&&600$k$ & 85.07/85.18 & 90.54 & 91.14 & 91.74 & 52.87 & 87.99 & 70.40 & 89.13 & 82.67  \\
&&&&&1$M$   & 85.55/85.39 & 91.09 & 91.14 & 92.43 & 54.80 & 87.01 & 69.31 & 89.71 & 82.94  \\
\midrule
\multirow{4}{*}{M4x3-SD}&\multirow{4}{*}{112$M$}&\multirow{4}{*}{(3648,768,12)}&\multirow{4}{*}{34}&\multirow{4}{*}{1.3x}&100$k$ & 82.44/82.82 & 88.82 & 90.86 & 90.71 & 48.84 & 84.80 & 66.07 & 87.80 & 80.35 \\
&&&&&300$k$ & 84.47/84.48 & 90.32 & 91.23 & 92.32 & 54.23 & 87.25 & 67.15 & 88.16 & 82.18 \\
&&&&&600$k$ & 85.07/85.03 & 90.81 & 91.32 & 92.43 & 55.04 & 88.48 & 68.95 & 88.75 & 82.88 \\
&&&&&1$M$   & 85.48/85.31 & 91.01 & 91.39 & 93.46 & 55.76 & 88.24 & 68.95 & 89.35 & 83.22  \\
\midrule
\multirow{4}{*}{M6x2-SD}&\multirow{4}{*}{112$M$}&\multirow{4}{*}{(3712,768,12)}&\multirow{4}{*}{33}&\multirow{4}{*}{1.4x}&100$k$ & 82.24/82.55 & 88.85 & 90.84 & 90.94 & 50.51 & 85.54 & 62.09 & 85.54 & 79.90 \\
&&&&&300$k$ & 83.72/83.93 & 90.35 & 91.14 & 92.32 & 52.34 & 87.75 & 64.26 & 86.30 & 81.34 \\
&&&&&600$k$ & 84.71/84.68 & 90.68 & 91.23 & 92.32 & 54.69 & 87.01 & 67.51 & 86.72 & 82.17 \\
&&&&&1$M$   & 84.93/85.04 & 90.98 & 91.34 & 92.55 & 55.27 & 87.25 & 67.87 & 86.98 & 82.47  \\
\midrule
\multirow{4}{*}{M5M3M2M2-SD}&\multirow{4}{*}{112$M$}&\multirow{4}{*}{(3584,768,12)}&\multirow{4}{*}{35}&\multirow{4}{*}{1.3x}&100$k$ & 82.36/82.87 & 89.11 & 90.86 & 91.06 & 47.62 & 85.54 & 68.59 & 88.37 & 80.71 \\
&&&&&300$k$ & 84.13/84.20 & 90.30 & 91.15 & 92.09 & 51.87 & 87.99 & 66.07 & 88.17 & 81.77 \\
&&&&&600$k$ & 85.07/84.96 & 90.85 & 91.35 & 92.32 & 53.64 & 89.22 & 68.23 & 88.56 & 82.69 \\
&&&&&1$M$   & 85.23/85.03 & 91.09 & 91.29 & 92.55 & 52.36 & 88.24 & 71.12 & 89.10 & 82.89 \\
\midrule
\multirow{4}{*}{M2M2M3M5-SD}&\multirow{4}{*}{112$M$}&\multirow{4}{*}{(3584,768,12)}&\multirow{4}{*}{35}&\multirow{4}{*}{1.3x}&100$k$ & 82.14/82.76 & 88.82 & 90.74 & 90.94 & 49.63 & 87.25 & 63.54 & 87.80 & 80.40 \\
&&&&&300$k$ & 84.30/84.47 & 90.44 & 91.07 & 92.32 & 51.16 & 86.76 & 65.70 & 88.44 & 81.63 \\
&&&&&600$k$ & 85.09/85.25 & 90.66 & 91.15 & 92.89 & 53.41 & 87.99 & 68.59 & 89.02 & 82.67 \\
&&&&&1$M$   & 85.47/85.54 & 90.83 & 91.13 & 92.89 & 53.48 & 88.24 & 68.23 & 89.54 & 82.82 \\
\bottomrule

% part3
\bottomrule
\multirow{4}{*}{M2x6}&\multirow{4}{*}{161$M$}&\multirow{4}{*}{(6144,768,12)}&\multirow{4}{*}{45}&\multirow{4}{*}{1.0x}&100$k$ & 82.76/82.96 & 89.91 & 90.66 & 90.83 & 50.49 & 85.29 & 64.26 & 87.45 & 80.51 \\
&&&&&300$k$ & 85.11/85.14 & 91.12 & 91.14 & 92.09 & 55.21 & 86.76 & 70.04 & 89.12 & 82.86 \\
&&&&&600$k$ & 85.96/85.93 & 91.78 & 91.32 & 93.12 & 58.80 & 87.99 & 70.76 & 89.67 & 83.93 \\
&&&&&1$M$   & 86.08/86.57 & 91.76 & 91.39 & 92.89 & 60.65 & 87.99 & 75.45 & 90.18 & 84.77 \\
\midrule
\multirow{4}{*}{M2x7}&\multirow{4}{*}{148$M$}&\multirow{4}{*}{(4480,768,12)}&\multirow{4}{*}{46}&\multirow{4}{*}{1.0x}&100$k$ & 82.20/82.46 & 89.36 & 90.63 & 90.94 & 48.19 & 85.54 & 66.43 & 88.04 & 80.42 \\
&&&&&300$k$ & 85.23/84.86 & 90.94 & 91.03 & 92.20 & 56.12 & 87.01 & 68.23 & 89.20 & 82.76 \\
&&&&&600$k$ & 85.89/85.94 & 91.31 & 91.30 & 92.20 & 57.01 & 88.73 & 71.84 & 89.66 & 83.76 \\
&&&&&1$M$   & 86.25/86.26 & 91.71 & 91.28 & 92.55 & 61.34 & 88.24 & 75.09 & 90.02 & 84.75 \\
\bottomrule

\end{tabular}
\addtolength{\tabcolsep}{2.2pt}
\end{sidewaystable}

\end{document}